\documentclass[10pt,final,a4paper]{article}

\usepackage{graphicx}
\usepackage{amsmath} 
\usepackage{subfigure}
\usepackage{multirow}
\usepackage{verbatim}
\usepackage{cite}
\usepackage[small,labelfont=bf]{caption}

\title{\LARGE \bf Bio-inspired friction switches: adaptive pulley systems}

\author{Konstantinos Dermitzakis and Juan Pablo Carbajal
\thanks{K. Dermitzakis is with the AILab, UZH, Switzerland
        {\tt\footnotesize dermitza@ifi.uzh.ch}.
				J.P. Carbajal is with the ELIS Department, UGhent, Belgium
				{\tt\footnotesize juanpablo.carbajal@ugent.be}.}%
}

\begin{document}

\maketitle
\thispagestyle{empty}
\pagestyle{empty}

\begin{abstract}

Frictional influences in tendon-driven robotic systems are generally unwanted, with efforts towards minimizing them where possible. In the human hand however, the tendon-pulley system is found to be frictional with a difference between high-loaded static post-eccentric and post-concentric force production of 9-12\% of the total output force. This difference can be directly attributed to tendon-pulley friction. Exploiting this phenomenon for robotic and prosthetic applications we can achieve a reduction of actuator size, weight and consequently energy consumption. In this study, we present the design of a bio-inspired friction switch. The adaptive pulley is designed to minimize the influence of frictional forces under low and medium-loading conditions and maximize it under high-loading conditions. This is achieved with a dual-material system that consists of a high-friction silicone substrate and low-friction polished steel pins. The system, designed to switch its frictional properties between the low-loaded and high-loaded conditions, is described and its behavior experimentally validated with respect to the number and spacing of pins. The results validate its intended behavior, making it a viable choice for robotic tendon-driven systems.

\end{abstract}

\section{Introduction}

Tendon-driven robotic systems are becoming increasingly favorable since they facilitate compliance through elasticity~\cite{2012_Humanoids_Lens} and lighweight end effectors~\cite{DeLuca20051809}. The use of tendons allows for designs that can incorporate compliant joints, adding safety to systems interacting with humans. Furthermore, tendon-driven systems maintain the end-effector inertias and loads independent of the actuator mass and location. In the design of these systems, significant effort is placed on minimizing effects introduced by tendon-pulley friction such as hysteresis, nonlinearities~\cite{Smagt_2009} and energetic losses~\cite{5308286, 4456748, lali2010}. This can be observed across most systems, and particularly in robotic hands, e.g. the Smart Hand \cite{biomechatronicDesign}, the Shadow Hand \cite{shadowhand} and others\cite{5308286, 5627148}. To minimize the tendon-pulley frictional influences, Teflon or Nylon coated cables and pulleys with ball-bearings are utilized.

However, in robotic applications, and particularly in prosthetic systems, additional counter-acting requirements need to be met, e.g. reduced power consumption and low overall device weight~\cite{5627148}. In these cases, designing the system to be tendon-driven alone is not enough. We argue that by studying the biomechanics of the human hand we can identify and exploit principles~\cite{Pfeifer16112007} that can be used to minimize the introduced trade-offs while still being practical for applications in robotics.

Investigating the tendon-pulley system of the human finger, post-eccentric and post-concentric\footnote{The terms \textsl{post-eccentric and post-concentric configuration} will be used here to denote a static configuration from which an eccentric or concentric contraction would follow if the loading force would be increased or decreased by a very small amount respectively.} forces differ by 9-12\% during high-load flexion of the interphalangeal joints. This difference can be directly attributed to tendon-pulley friction~\cite{Schweizer200363} and is beneficial as it enables a higher fingertip output force production during a post-eccentric configuration. The benefit provided by tendon-pulley friction can be clearly identified if we note the different kinds of activities in the context of eccentric and concentric configurations. Matsumoto et al.~\cite{conceptICF} recorded a life-log of a healthy person and analyzed it based on the ICF\footnote{International Classification of Functioning, Disability and Health, also known as ICF, is a classification of the health components of functioning and disability.}. There is a number of activities that are of eccentric nature concerning the fingers: (1) carrying in the hands, (2) pulling, (3) lifting and (4) putting down objects. Considering that these activities compose the majority of daily performed actions of a healthy person, summing over $60\%$, the benefit of friction in eccentric configurations is apparent.


Based on these facts, we question the ongoing principle of minimizing friction in robotic tendon-driven systems. Friction in the tendon-pulley system of such devices, and in particular robotic hands, can be of great benefit for high-loaded post-eccentric configurations. At these configurations, the mechanical device can be assumed to be operating at its maximum load. With the introduction of a frictional tendon-pulley system the device can have an extra fraction of output force at no additional motor cost. Alternatively, the motor weight can be scaled down (assuming a linear relationship of mass-torque~\cite{Dermitzakis2011}) by that same fraction. This reduction is possible while maintaining a comparable force output to a device with no frictional tendon-pulley system. The latter scenario is particularly interesting for prosthetic devices, as one of the functional factors of device rejection is its weight~\cite{pmid18050007, Carrozza2006}.

As such, we have designed a mechanical pulley system that switches between a low and a high coefficient of friction based on the normal force the tendon is holding. By utilizing such a transition, we can ensure that the frictional disadvantages of the tendon-pulley system are minimized during low to medium loading, while at high loading, post-eccentric frictional forces are being used to the advantage of the actuators of the intended system. Such a design can be directly applied to tendon-driven robotic (and) prosthetic hands but also to tendon-driven robotic devices that operate under similar conditions.

\section{Pulley design}

\subsection{Adaptive friction pulley}
\label{sec:design}

\begin{figure}
\centering
\includegraphics[width=0.9\textwidth]{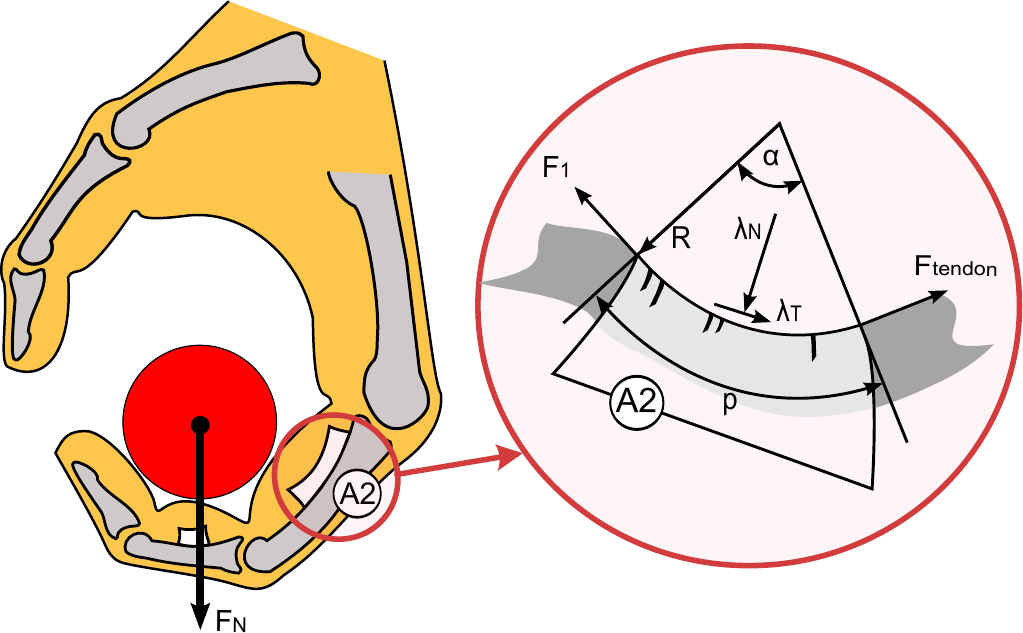}
\caption{Left: The A2 pulley in the human finger during maximum finger contraction. Right: In this configuration the A2 pulley can be modeled as a capstan segment.}
\label{fig_pulley}
\end{figure}

The adaptive pulley has been designed after the A2 pulley as it has been shown to be the main contributor of friction in the human finger~\cite{Roloff16488229, Schweizer2008}. Furthermore, it is modeled at maximum finger contraction (Fig. \ref{fig_pulley}). In that posture and assuming maximum loading forces, pulley-loading is maximum with the pulley taking an elliptical shape that can be approximated as a circle segment. A thin grooved layer of silicone with a high coefficient of friction is placed on top of a rigid ABS shell. This layer serves as the high-frictional substrate that the \textsl{tendon} will come in contact with under high-loaded configurations. On top of the silicone layer, steel pins are suspended in the grooves by means of side supports with the bottom rigid material. The side supports further help the silicone layer to stay in place when in contact with the tendon under motion.

\begin{figure}
\centering
\includegraphics[width=0.9\textwidth]{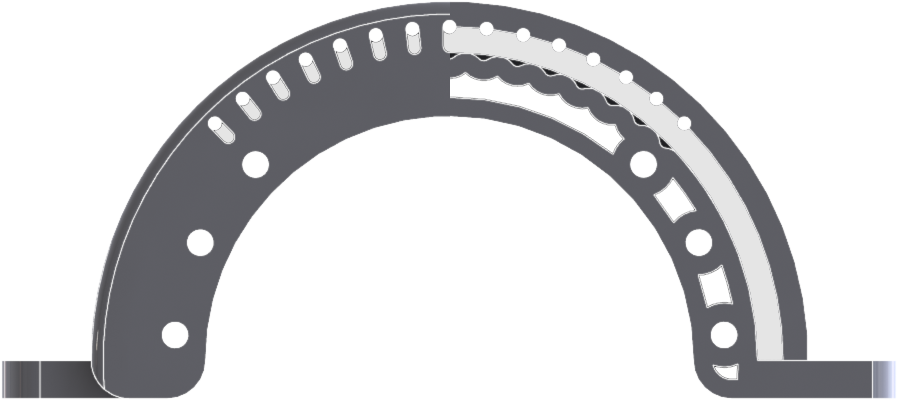}
\caption{The friction switch. It consists of a soft substrate, shown in light gray and grooves ($D_g=1mm$) for 15 pins over an angle of $89.45^\circ$. The side support assures that the pins travel axially to the pulley under load.}
\label{fig_device}
\end{figure}

Under low-loaded configurations, the \textsl{tendon} will only come in contact with the pins, and thus the frictional forces will be minimal due to the low friction coefficient between the two materials in contact. As the external load increases, the pins sink in the silicone, which in turn sinks in the grooves of the hard shell. This will gradually increase friction up to a point where the pins will be fully sunk in the silicon, i.e. the high-frictional substrate. At this point, the frictional influence of the silicone will be at its maximum. There will exist a threshold external load at which point there will be a transition between the frictional properties of the pins and the frictional property of the silicone. As some of the pulley surface is actually occupied by the pins, the maximum frictional force observed will always be lower than the frictional force developed using only the silicone substrate.

The pins are suspended in place with side supports that only allow for axial motion to the circular segment. There are 15 grooves, with an inter-groove angle of $5.96^\circ$, for a total segment angle of $89.45^\circ$. The radius of the grooves on both the hard shell and the silicone is $0.5mm$. The silicone substrate is $2mm$ thick and is a silicone elastomer Sylgard 184, $50$ Shore A. The pins used are tin electroplated polished steel and are of diameter $D_{pin}=(0.94 \pm0.01)mm$. The wire used as a tendon is a Carl Stahl stainless steel wire rope model $U 8199512$. It is coated with Polyamid 12, has a radius of $1.2mm$ and a minimum breaking load of $850N$.

\section{Experimental design}
\label{sec:setup}

\begin{figure}
\centering
\includegraphics[width=0.8\textwidth]{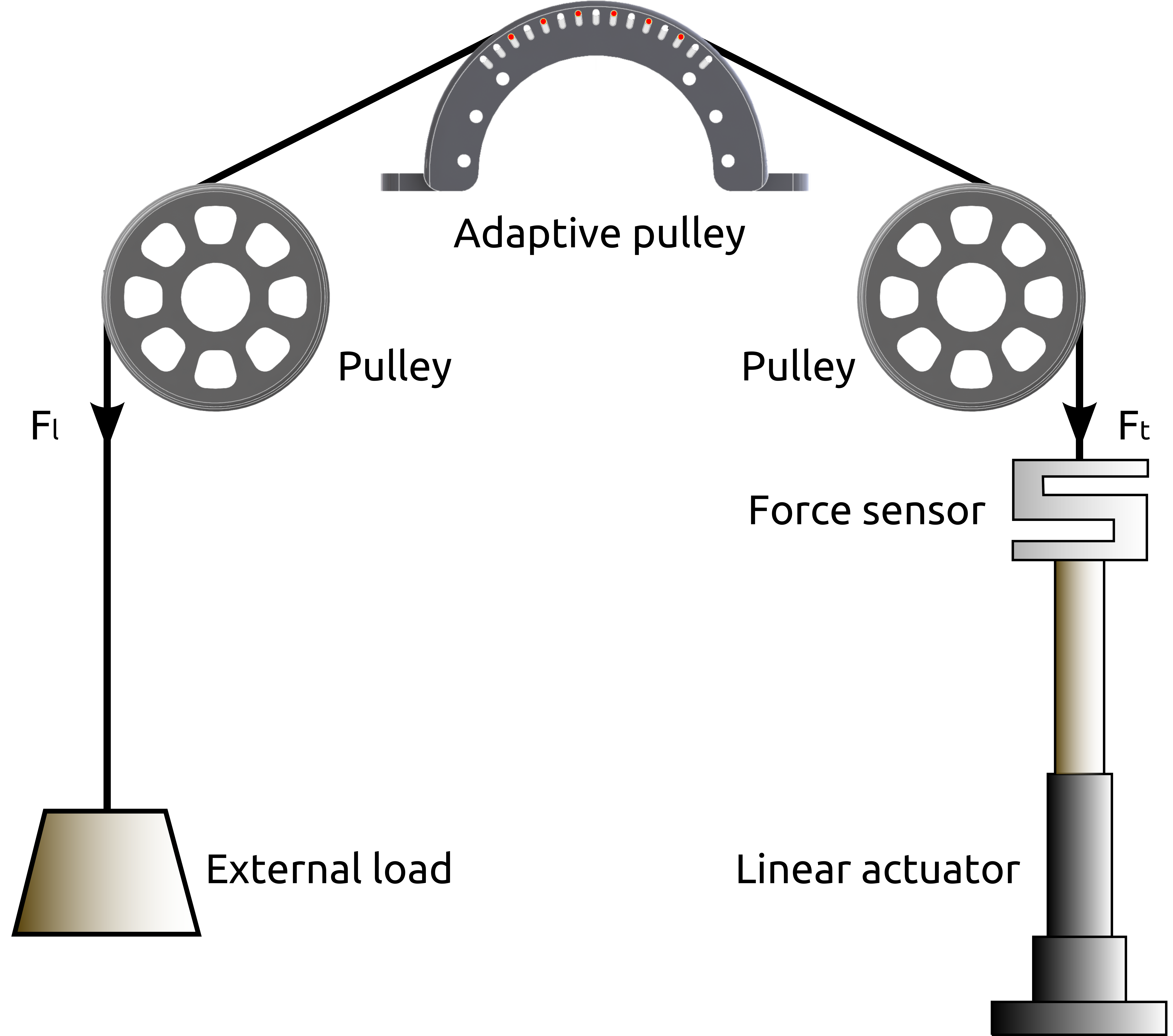}
\caption{The experimental setup consists of the friction switch (center) and two external pulleys that are used to route the tendon to the external load (left) and the linear actuator (right).}
\label{fig_setup}
\end{figure}

A number of experimental setups have been used to measure the system friction of a biological tendon-pulley, with the most prominent being that of Uchiyama et al. \cite{UchiyamaGlidingResistance}. We designed a similar setup for testing the adaptive pulley system, as seen in Fig. \ref{fig_setup}. It consists of a fixed two-pulley system, a linear actuator, a set of weights and a load cell. The friction switch is mounted on a screw gear system in the center of the setup, between the two pulleys. The pulleys are 3D-printed using a Dimension Elite 3D printer, have a radius of $r_{p}=22mm$ and encase industrial ABEC-5 $608z$ ball-bearings. They are used for guiding the tendon from the loads, over the friction switch and into the force sensor. The force sensor is a Me-systeme KD-40s, rated for a maximum load of $\pm$100N. It is attached to the linear actuator using a two-hook system that allows vertical rotation between the sensor and the actuator. The linear actuator used is a Firgelli Automations FA-PO-35-12-6, with a maximum force of $155.58N$ and an actuating speed of $50mm/s$. For the experiments performed, eleven calibrated weights were used; their values are shown in Table \ref{tab:weights}. The weights were attached to the tendon with a safety hook that weighs $(62.53 \pm0.01)\times 10^{-3}\; kg$.

\begin{table}
\renewcommand{\arraystretch}{1.3}
\caption{Calibrated weights and their respective actuating speeds}
\label{tab:weights}
\centering
\begin{tabular}{c|c}
\hline
Weight (grams) & Actuator speed (\%)\\
\hline \hline
$251.2 \pm0.1$ & $47\%$\\
\hline
$503.6 \pm0.1$ & $48\%$\\
\hline
$1003.2 \pm0.1$ & $49\%$\\
\hline
$1503.9 \pm0.1$ & $50\%$\\
\hline
$2005 \pm0.1$ & $51\%$\\
\hline
$2505.4 \pm0.1$ & $52\%$\\
\hline
$3007.2 \pm0.1$ & $53\%$\\
\hline
$3508.6 \pm0.2$ & $54\%$\\
\hline
$4007.8 \pm0.2$ & $55\%$\\
\hline
$4504.5 \pm0.2$ & $56\%$\\
\hline
$5002.5 \pm0.2$ & $57\%$\\
\hline
\end{tabular}
\end{table}

The linear actuator is PC-controlled via a TITech SH2Tiny microcontroller and a Pololu MD03A motor driver. The force sensor has been calibrated with a GSV-11 differential amplifier for a full 0-5V scale that corresponds to 0-5Kg load. Data is sampled and recorded in real time with a sampling frequency of 200Hz and includes (1) forces (2) actuator positions and (3) actuator power consumption. The control of the actuator is open-loop PWM, and corresponds to 47\% of the full actuator speed for the initial weight, and a 1\% increment for every subsequent weight, with a final actuator speed of 58\% at maximum load.

To obtain the static frictional forces, the actuator is controlled to move through the full actuator stroke of $15.24cm$ by first extending and subsequently contracting with the loads attached. The procedure is as follows: The actuator is turned on at the corresponding load speed for $500ms$, and is then turned off for $3000ms$. This is repeated for the full actuator stroke, first by releasing the load (eccentric) and consequently for contracting with the load (concentric). As the control is open-loop, and due to the attached loads, there is an assymetry between the repetitions for extending and contracting, with respectively 13 and 30 repetitions per load in average.

\begin{figure}
\centering
\subfigure[Six pins, single spacing configuration]{\includegraphics[width=0.75\linewidth]{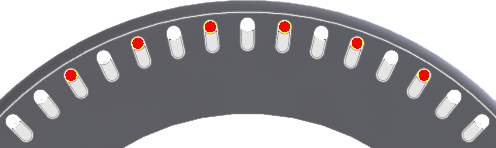}
\label{sixPins}}
\vfil
\subfigure[Three pins, double spacing configuration]{\includegraphics[width=0.75\linewidth]{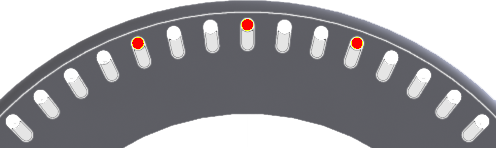}
\label{threePinsDouble}}
\vfil
\subfigure[Three Pins, single spacing configuration]{\includegraphics[width=0.75\linewidth]{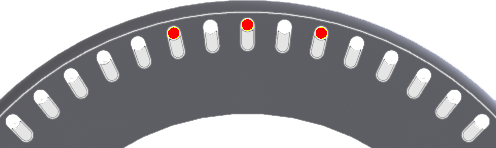}
\label{threePinsSingle}}
\caption{The three different experimental pin configurations with the pin positions marked in red. From top to bottom, six single-spaced pins, three double-spaced pins and three single-spaced pins.}
\label{fig_pins}
\end{figure}

To identify the contribution of the silicone substrate and the pins alone, the system was tested under all loads without any pins attached to provide the characteristic behavior of the silicone. Further, a \textsl{proxy} silicone substrate was created from ABS plastic, and all three pin configurations were tested under all loads to identify the characteristic behavior of three pin configurations. These pin configurations serve to verify the intended behavior of the system, and are shown in Fig. \ref{fig_pins}. The first configuration uses six pins with a single groove gap between them. The inter-pin angle is $5.96^\circ$ and the maximum spanned pin contact angle is $59.6^\circ$. The second configuration uses three pins with two groove gaps between them. The inter-pin angle is $11.92^\circ$ and the maximum spanned pin contact angle is $35.76^\circ$. The last configuration uses three single-spaced pins, with an inter-pin distance of $5.96^\circ$ and a maximum spanned pin contact angle of $23.84^\circ$. For all configurations, the respective contact angle of the \textsl{tendon} with the friction switch is fixed at $63.89^\circ$.

\section{Results}
\label{sec:results}

To extract the post-eccentric and post-concentric frictional forces, data is post-processed offline in MATLAB (2009, The MathWorks, Inc.). All repetitions for each condition and load respectively are first super-imposed and then manually segmented. The segmented data is then used to calculate the means and standard deviations of forces involved for each respective load and loading condition. For the absolute frictional force magnitude $F_{fr}=\left|(F_{e}-F_{c})/2\right|$ plots, the standard deviation is calculated as $s_{fr}=(s_{e}+s_{c})/2$ and is used to display a region around the plotted data for clarity.

\subsection{Empirical friction switching model}
\label{sec:model}

In order to describe the phenomenological behavior of the friction switch, we have used a frictional two-link, one joint, finger tendon-pulley model that has been show to correspond well with physiological data~\cite{Dermitzakis-2012-ID22}. The frictional tendon-pulley interaction is modeled using the capstan friction equation~\cite{levin1991}. In a scenario where the loading force $F_{l}$ is fixed, the system equation and the Capstan equations are:

\begin{eqnarray}
	F_{t}^{e}  &=& F_{l} \cdot e^{-\mu \cdot \alpha} \label{eq:tendonecc} \\
	F_{t}^{c} &=& F_{l} \cdot e^{ \mu \cdot \alpha} \label{eq:tendonconc}
\end{eqnarray}

\begin{figure}
\centering
\includegraphics[width=0.9\linewidth]{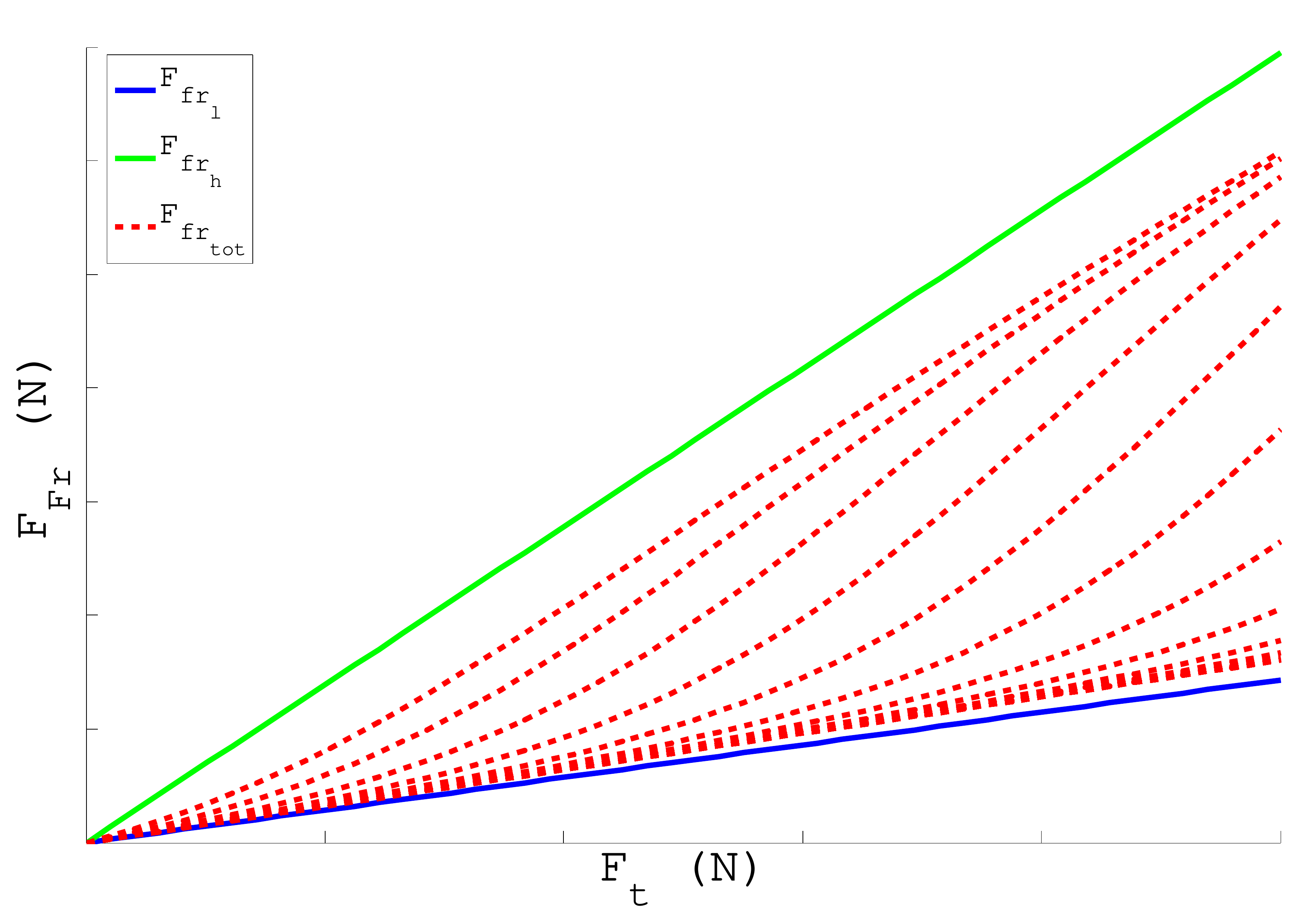}
\caption{The intended behavior of the adaptive pulley system. Given low-friction (blue) and high-friction (green) materials, the friction of the system (red) should transition between the two after some threshold external normal force $F_N$.}
\label{fig_model}
\end{figure}

Where $F_{t}$ is the tendon force, or in this study, the forces observed by the force sensor. Our objective is to describe the behavior of a system where there is a smooth transition between two materials with different frictional coefficients but otherwise identical $F_{l}$ and $\alpha$. We can model such a transition by using a sigmoid transfer function of the form $S(F_{l})=1/1+e^{-(F_{w}-F{thr})}$, where $F_{r}$ is the $F_{l}$ range of forces where the transfer is happening and $F_{thr}$ defines the threshold (middle) point of the transfer function. By combining the frictional force equations of each material we end up with:

\begin{eqnarray}
	F_{fr_{tot}} &=& F_{fr_{l}}(1 - 1/(1+e^{-(F_{r}-F_{thr})}))\nonumber\\
	&&+ (F_{fr_{h}}/(1+e^{-(F_{r}-F_{thr})})
\end{eqnarray}

Where $F_{fr_l}, F_{fr_h}$ are the frictional forces of the low friction and high friction materials respectively. Such a function makes the assumption that the transition happens with both material having full contact with the tendon at the same time. As this cannot happen with a physical system we introduce the weighing factor $w$:

\begin{eqnarray}
	F_{fr_{tot}} &=& (1+w)F_{fr_{l}}(1 - 1/(1+e^{-(F_{r}-F_{thr})}))\nonumber\\
	&& + (1-w)(F_{fr_{h}}/(1+e^{-(F_{r}-F_{thr})}) \label{eq:4}
\end{eqnarray}

The weighing factor is used to simulate the material distribution over the pulley surface, under the assumption that $w \leq 0.5$. The model behavior for a varying $F_{thr}$ can be seen in Fig. \ref{fig_model}. For fitting the model to the experimental data recorded, all configurations share the following parameters: $\alpha=63.89^\circ$, $\mu_{p}=0.05$, $\mu_{s}=0.24$. Further, $w=0.2$ for the six pin and $w=0.1$ for both three pin configurations.

\subsection{Silicone and pin characteristic behavior}

\begin{figure}
\centering
\includegraphics[width=\linewidth]{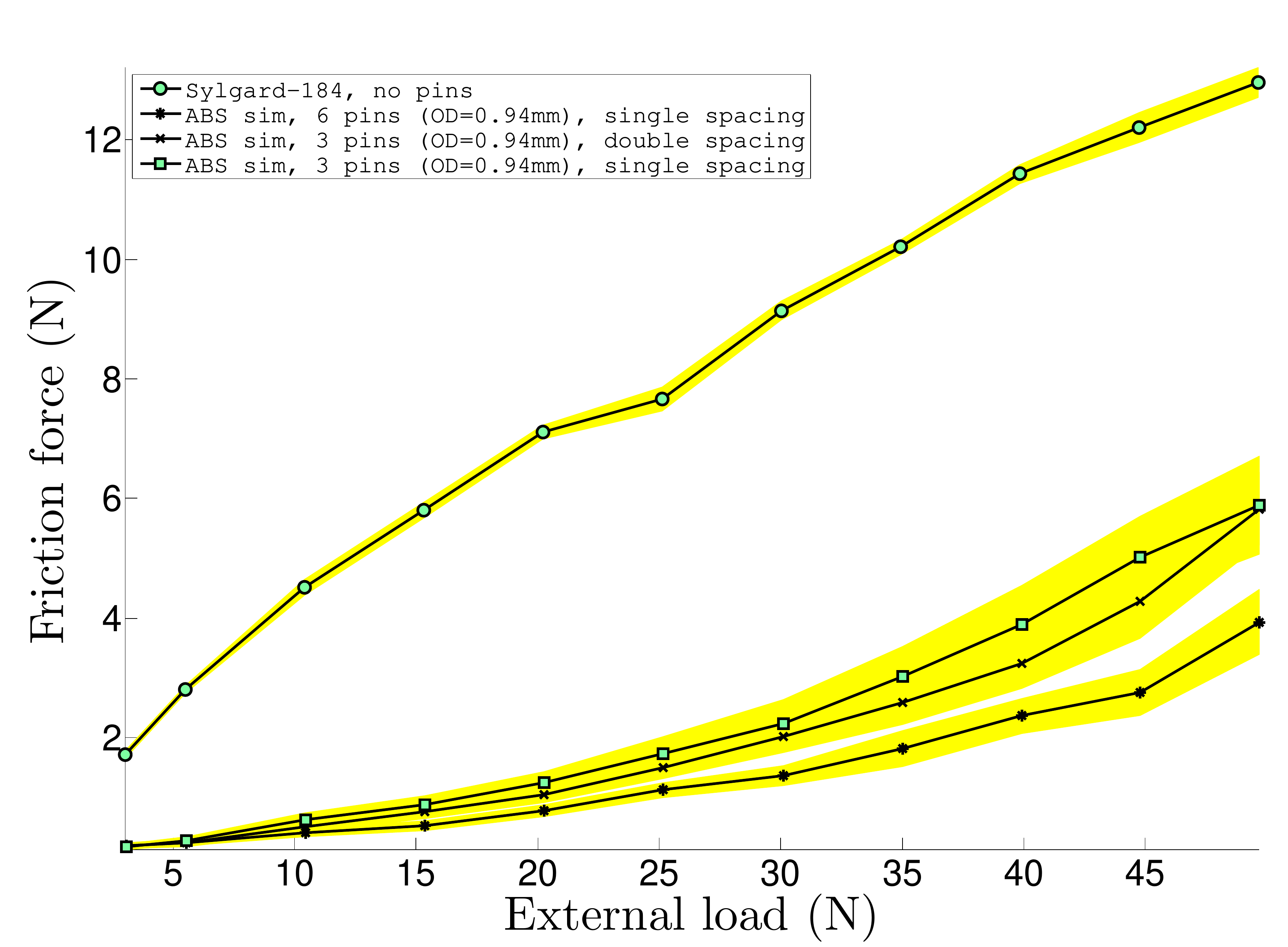}
\caption{The individual characteristic frictional force behaviors of the silicone substrate and all three pin configurations. The yellow regions denote the standard deviations of the absolute frictional forces and are displayed for clarity.}
\label{fig_siliconPins}
\end{figure}

The characteristic behaviors of the silicone substrate and pins can be seen in Fig. \ref{fig_siliconPins}. The silicone has a relatively linear behavior against external loads, with a maximum frictional force of $F_{fr_{sil}^{max}}=12.97N$, with a slight curvature observed. The six pin configuration produces a maximum force of $F_{fr_{six}^{max}}=3.935N$. The three double-spaced pin configuration produces a maximum force of $F_{fr_{thrd}^{max}}=5.829N$. Finally, the three single-spaced pin configuration produces a maximum force of $F_{fr_{thrs}^{max}}=5.893N$. All pin configurations display a non-linear trend of frictional forces based on the external load.

\subsection{Pin configurations}

\begin{figure}
\centering
\subfigure[Six pins, single spacing configuration]{\includegraphics[width=0.6\linewidth]{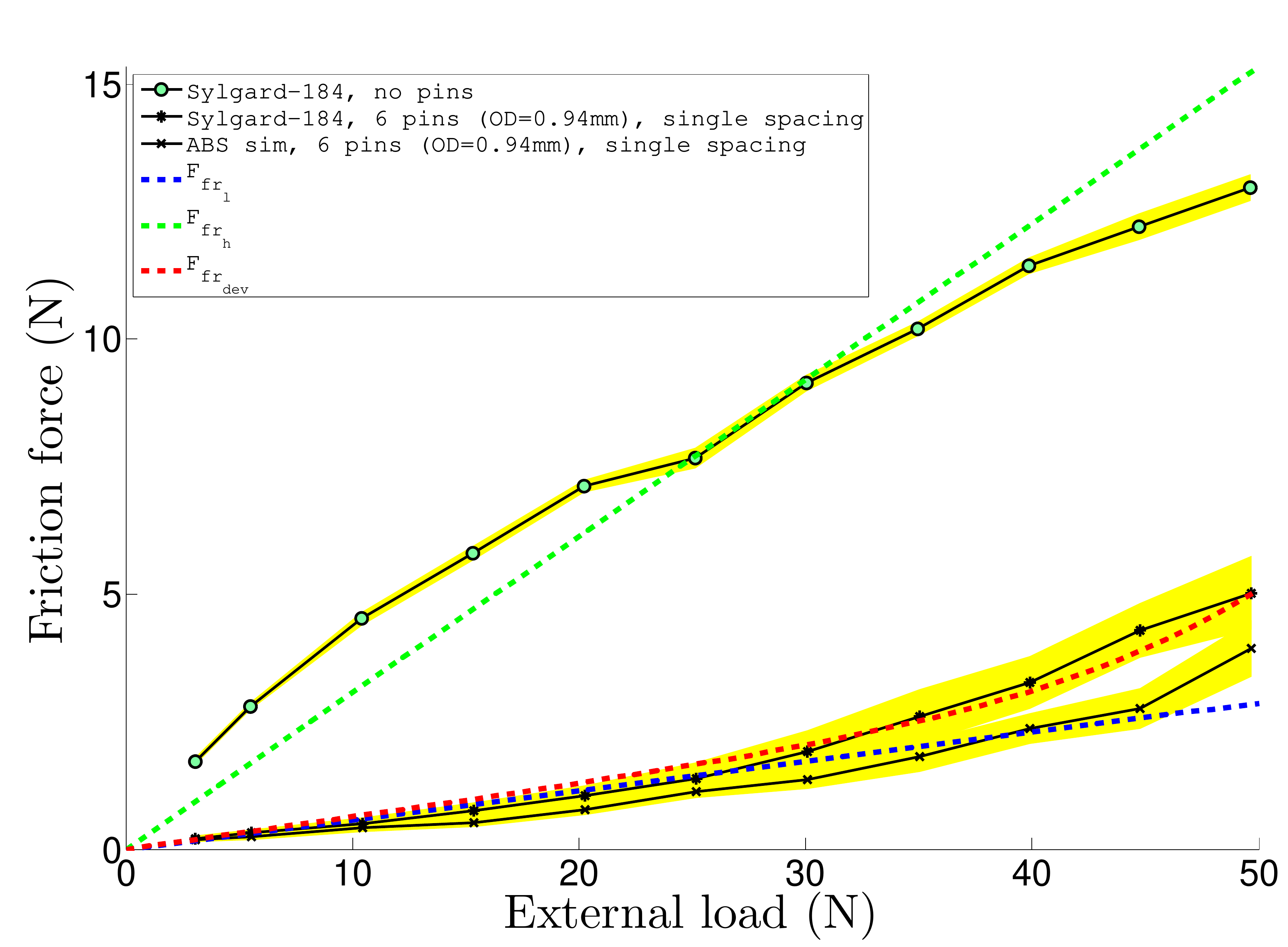}
\label{sixPinsR}}
\vfil
\subfigure[Three pins, double spacing configuration]{\includegraphics[width=0.6\linewidth]{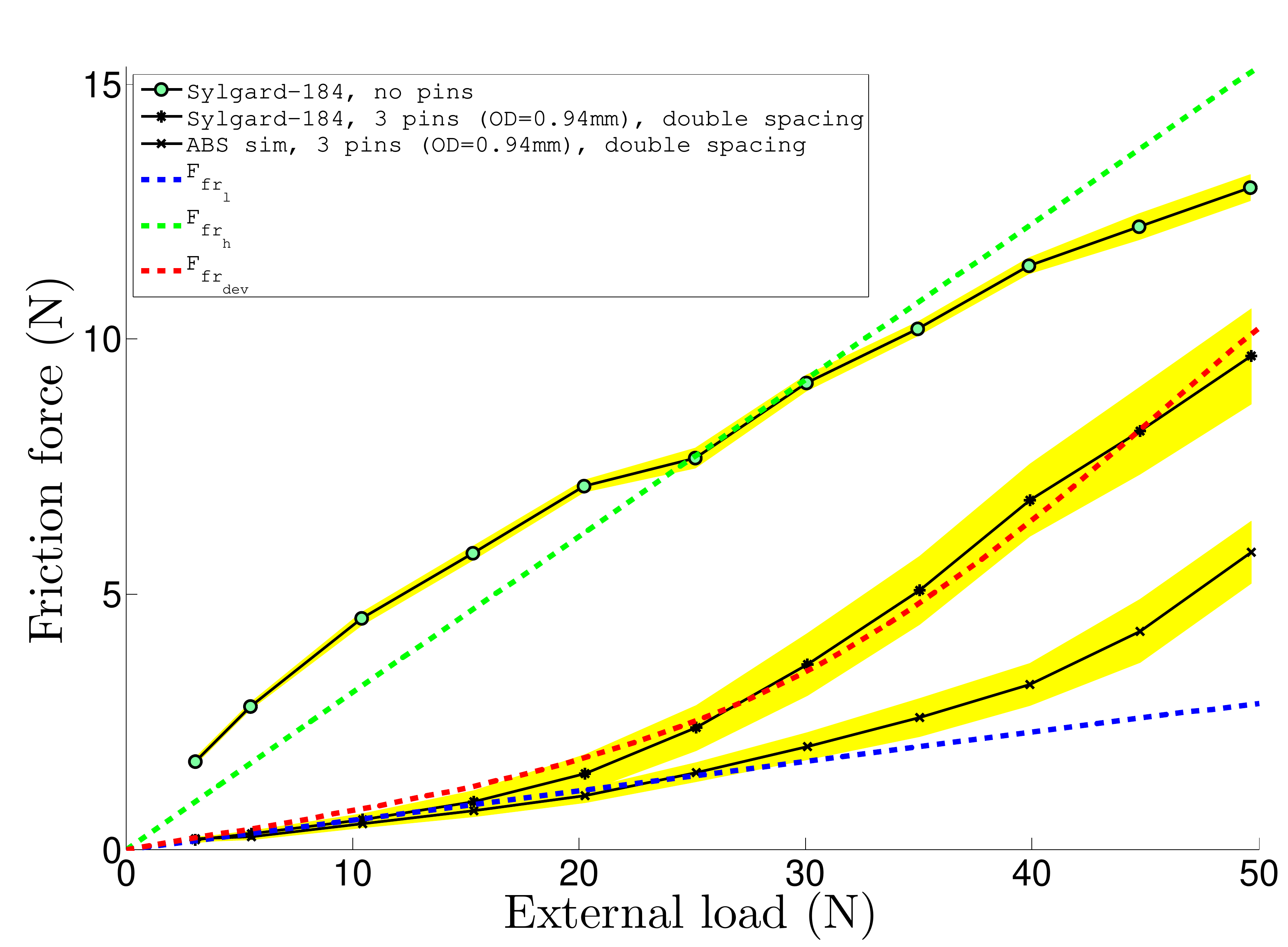}
\label{threePinsDoubleR}}
\vfil
\subfigure[Three Pins, single spacing configuration]{\includegraphics[width=0.6\linewidth]{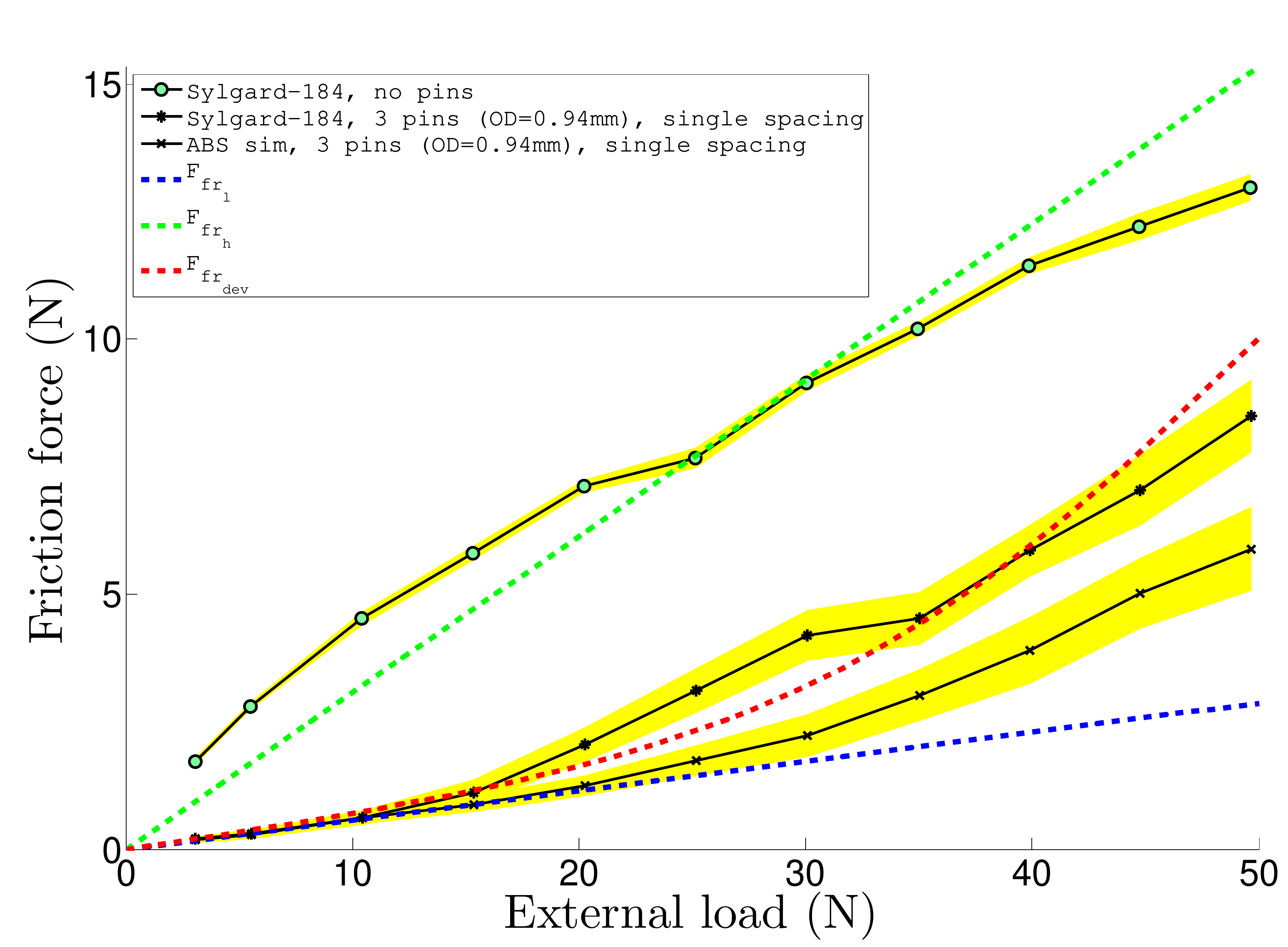}
\label{threePinsSingleR}}
\caption{The results of the three different experimental pin configurations. From top to bottom, six single-spaced pin, three double-spaced pin and three single-spaced pin configurations. The continuous lines denote experimental data and the segmented lines the fitted model. The yellow regions denote the standard deviations of the absolute frictional forces and are displayed for clarity.}
\label{fig_pinsR}
\end{figure}

\paragraph{\textbf{Six pins, single spacing, Fig. \ref{sixPinsR}}} In the six-pin, single spacing configuration, the system behaves similarly both with and without the silicone substrate, with a maximum friction force of $F_{fr_{dev}^{max}}=5.097N$. The model parameters used for the fitting are $F_{thr} = 5.7N$ and $F_r=\left\{-2,3.5\right\}N$.

\paragraph{\textbf{Three pins, double spacing, Fig. \ref{threePinsDoubleR}}} The three-pin, double spacing configuration clearly shows the friction transition effect between the pins and the substrate. The model parameters used for the fitting are $F_{thr} = 4.3N$ and $F_r=\left\{0, 5.5\right\}N$. Frictional forces behave as if only the pins are present until approximately $15.31N$, where the adaptive pulley transitions into a high-friction mode, with a maximum friction force of $F_{fr_{dev}^{max}}=9.667N$.

\paragraph{\textbf{Three pins, single spacing, Fig. \ref{threePinsSingleR}}} The three-pin, single spacing configuration displays a transition effect that lies in between the two other configurations, with a maximum friction force of $F_{fr_{dev}^{max}}=8.495N$. The model parameters used for the fitting are $F_{thr} = 4.3N$ and $F_r=\left\{0, 4.9\right\}N$.

\section{Discussion}

\subsection{Silicone and pin characteristic behavior}

The observed curvature of the silicone substrate in Fig. \ref{fig_siliconPins} hints towards the existence of a maximum substrate frictional force at some load. This is also true on a physical basis, as frictional forces of any material are upper-bounded; they cannot infinitely grow. However, larger external loads would be required to validate whether what we are observing is the upper frictional force limit of the silicone substrate in this particular case.

The pin behavior is peculiar in two aspects: (1) frictional forces of all pin configurations are non-linear, with a tendency to grow over larger external loads and (2) as the number of pins increases, friction decreases. The non-linear behavior of the pins can be explained as an effect of the Polyamide coating of the steel cable. The coating thickness is $0.25mm$ and as such at a scale comparable to the pins having a radius of $r_{pin}=0.47mm$. As both the pins and steel core of the \textsl{tendon} are rigid, we can expect deformations happening under load to the polymer coating. By increasing the external load, the deformations will be larger, increasing the frictional forces against the pins. This results in the observed non-linear effect, with visible polymer surface deformation with a microscope (20x magnification) under extreme loads.

Furthermore, with an increased number of pins, it forms a smooth shape under tension. As the number of pins is reduced, the \textsl{tendon} forms into disjointed linear segments with discontinuities at the pin locations. These segments account for increased friction, as both the \textsl{tendon} and its coating have to stress and deform over the pins to overcome the discontinuities of the formed linear segments, thereby resulting in increased observed frictional forces. This is also the reason for the model not precicely matching the pin behavior; the experimental observations not only contain the frictional characteristics of the pins but also the behavior of the tendon coating. In contrast, the tendon coating has a minimal influence on the frictional forces observed on the silicone; the substrate allows for a much greater deformation in comparison to the tendon coating, reducing its effects.

\subsection{Friction switch behavior}

The observed behavior of the friction switch in the six-pin configuration suggests that the forces involved are not sufficient in producing the intended pin-sinking switching effect, with the \textsl{tendon} not making sufficient contact with the silicone substrate. In contrast, the three-pin, double spacing configuration clearly shows the friction transition effect between the pins and the substrate. Of course it is not expected that the full frictional effect of the silicone substrate will be present, as detailed in Section \ref{sec:model}, however at the maximum tested load, the frictional difference between the pin-substrate and substrate configurations is only 27\%. Finally, the three-pin, single spacing configuration displays a transition effect that lies in between the two other configurations. In all cases, larger forces would be required to show the full transition effects between the two materials for all pin configurations.

Based on the experimental results, the adaptive pulley system behaves as intended, showing a transition between low and high friction that is clearly dependent on both the number and spacing of the pins. Pins spaced further apart lower the threshold force where the switch between low and high frictional forces occurs. In addition, the location of where the pins are placed is also relevant to how and where the transition threshold appears, with pins placed in locations where the major tangential forces are involved increasing the external force transition threshold.

In addition to the transition effects, the pins serve a second function: they maintain the substrate in place and minimize excessive silicone deformations. During experimentation with only the substrate, large shearing deformations were observed with increasing loads. This is one of the effects that contributes to the large frictional forces present. Due to the design of the system, and with the pins in place, the substrate is essentially constrained between each pin pair. This segmentation allows the substrate to only deform between each respective pin pair and thus large, complete surface deformations are prevented. This is an additional reason of not reaching the full frictional forces present with only the substrate; the system limits the mechanical behavior of the substrate and in turn modifies its frictional characteristics.

\subsection{System optimizations}

Even though the intended behavior of the friction switch is verified, we still need to gain a fundamental understanding of the parameters affecting the frictional switching properties: (1) transition range, (2) force spread and (3) transition threshold. These are (1) pin diameter, (2) pin material, (3) substrate groove geometry, and (4) substrate material. We hypothesize that the pin/groove diameter can affect not only the transition threshold but also the transition range. Exploring alternative pin material can be useful for further minimizing the frictional properties of the system at low to medium loads. Different substrate materials will play a significant role in the maximum obtainable frictional forces at high loads and to an extent, also contribute to the transition threshold and range. Finally, it would be worthwhile to also incorporate the property of directional friction, such that the force output disadvantage at post-concentric configurations is reduced.


\section{Conclusion}

We have presented an adaptive pulley system that displays friction switching at particular external loading thresholds based on the number of pins used. This friction switch system can be exploited in any tendon-driven device that requires asymmetric maximum force production, and in particular hands, as it can reduce the actuator size required in achieving a specific maximum force. In turn it can reduce the total device weight which is a primary functional factor for device acceptance. This can be achieved by scaling the actuators for the lowest maximum force production required, with the friction switch providing the asymmetric maximum through its frictional properties. Further, this assymetry can be minimized to negligible levels at lower loads, in order to counteract unwanted frictional effects, if so required.

\section*{Acknowledgments}

The authors would like to thank Vu Quy Hung, Naveen Kuppuswamy and Prof. Dr. Rolf Pfeifer from the AI Lab, UZH for their valuable input. This research was supported by the SNSF through the National Centre of Competence in Research Robotics (NCCR Robotics).

\bibliographystyle{unsrt}
\bibliography{IEEEabrv,thumbbib}

\begin{thebibliography}{10}

\bibitem{2012_Humanoids_Lens}
Thomas Lens, Jérôme Kirchhoff, and Oskar von Stryk.
\newblock Dynamic modeling of elastic tendon actuators with tendon slackening.
\newblock In {\em Proceedings of the IEEE/RAS International Conference on
  Humanoid Robots (HUMANOIDS)}, page to appear, 2012.

\bibitem{DeLuca20051809}
Alessandro~De Luca, Bruno Siciliano, and Loredana Zollo.
\newblock Pd control with on-line gravity compensation for robots with elastic
  joints: Theory and experiments.
\newblock {\em Automatica}, 41(10):1809 -- 1819, 2005.

\bibitem{Smagt_2009}
Patrick van~der Smagt, Markus Grebenstein, Holger Urbanek, Nadine Fligge,
  Michael Strohmayr, Georg Stillfried, Jonathon Parrish, and Agneta Gustus.
\newblock Robotics of human movements.
\newblock {\em Journal of physiology, Paris}, 103(3-5):119--32, 2009.

\bibitem{5308286}
S.A. Dalley, T.E. Wiste, T.J. Withrow, and M.~Goldfarb.
\newblock Design of a multifunctional anthropomorphic prosthetic hand with
  extrinsic actuation.
\newblock {\em Mechatronics, IEEE/ASME Transactions on}, 14(6):699 --706, dec.
  2009.

\bibitem{4456748}
K.B. Fite, T.J. Withrow, Xiangrong Shen, K.W. Wait, J.E. Mitchell, and
  M.~Goldfarb.
\newblock A gas-actuated anthropomorphic prosthesis for transhumeral amputees.
\newblock {\em Robotics, IEEE Transactions on}, 24(1):159 --169, feb. 2008.

\bibitem{lali2010}
T.~Laliberté, Baril M., Guay F., and Gosselin C.
\newblock Towards the design of a prosthetic underactuated hand.
\newblock {\em Mechanical Sciences Journal}, 2:19--26, 2010.

\bibitem{biomechatronicDesign}
Loredana Zollo, Stefano Roccella, Eugenio Guglielmelli, M.~Chiara Carrozza, and
  Paolo Dario.
\newblock Biomechatronic design and control of an anthropomorphic artificial
  hand for prosthetic and robotic applications.
\newblock {\em IEEE/ASME T}, 12:4, 2007.

\bibitem{shadowhand}
{Shadow Robot Company}.
\newblock The shadow dextrous hand.
\newblock Online.

\bibitem{5627148}
M.~Controzzi, C.~Cipriani, B.~Jehenne, M.~Donati, and M.C. Carrozza.
\newblock Bio-inspired mechanical design of a tendon-driven dexterous
  prosthetic hand.
\newblock In {\em Engineering in Medicine and Biology Society (EMBC), 2010
  Annual International Conference of the IEEE}, pages 499 --502, 31 2010-sept.
  4 2010.

\bibitem{Pfeifer16112007}
Rolf Pfeifer, Max Lungarella, and Fumiya Iida.
\newblock Self-organization, embodiment, and biologically inspired robotics.
\newblock {\em Science}, 318(5853):1088--1093, 2007.

\bibitem{Schweizer200363}
A.~Schweizer, O.~Frank, P.~E. Ochsner, and H.~A.~C. Jacob.
\newblock Friction between human finger flexor tendons and pulleys at high
  loads.
\newblock {\em Journal of Biomechanics}, 36(1):63 -- 71, 2003.

\bibitem{conceptICF}
Yoshio Matsumoto, Yoshifumi Nishida, Yoichi Motomura, and Yayoi Okawa.
\newblock A concept of needs-oriented design and evaluation of assistive robots
  based on {ICF}.
\newblock {\em IEEE International Conference on Rehabilitation Robotics}, 2011.

\bibitem{Dermitzakis2011}
Konstantinos Dermitzakis, Juan~Pablo Carbajal, and James~H. Marden.
\newblock {Scaling Laws in Robotics}.
\newblock {\em Procedia Computer Science}, 7:250--252, January 2011.

\bibitem{pmid18050007}
C.~Pylatiuk, S.~Schulz, and L.~Doderlein.
\newblock {{R}esults of an {I}nternet survey of myoelectric prosthetic hand
  users}.
\newblock {\em Prosthet Orthot Int}, 31(4):362--370, Dec 2007.

\bibitem{Carrozza2006}
MC~Carrozza, G.~Cappiello, S.~Micera, BB~Edin, L.~Beccai, and C.~Cipriani.
\newblock {Design of a cybernetic hand for perception and action}.
\newblock {\em Biological Cybernetics}, 95(6):629--644, 2006.

\bibitem{Roloff16488229}
I.~Roloff, V.~R. Schoffl, L.~Vigouroux, and F.~Quaine.
\newblock {{B}iomechanical model for the determination of the forces acting on
  the finger pulley system}.
\newblock {\em J Biomech}, 39(5):915--923, 2006.

\bibitem{Schweizer2008}
Andreas Schweizer.
\newblock Biomechanics of the interaction of finger flexor tendons and pulleys
  in rock climbing.
\newblock {\em Sports Technology}, 1(6):249--256, 2008.

\bibitem{UchiyamaGlidingResistance}
S.~Uchiyama, Peter~C. Amadio, and K.N. An.
\newblock Gliding resistance of extrasynovial and intrasynovial tendons through
  the {A2} pulley.
\newblock {\em The Journal of Bone and Joint Surgery}, 79-A:219--224, 1997.

\bibitem{Dermitzakis-2012-ID22}
Konstantinos Dermitzakis, Marco~Roberto Morales, and Andreas Schweizer.
\newblock Modeling the frictional interaction in the tendon-pulley system of
  the human finger for use in robotics.
\newblock {\em Artif Life}, Nov 2012.

\bibitem{levin1991}
E.~{Levin}.
\newblock {Friction experiments with a capstan}.
\newblock {\em American Journal of Physics}, 59:80--84, January 1991.

\end{thebibliography}

\end{document}